\documentclass{svproc}
\usepackage{url}

\usepackage{hyperref}
\usepackage{url}
\usepackage[utf8]{inputenc}
\usepackage{graphicx}
\usepackage{amsmath} 
\usepackage{mathtools, nccmath}
\usepackage[
    style=numeric,
    sorting=none
]{biblatex}
\usepackage{adjustbox}
\usepackage{fancyhdr}
\usepackage{caption}

\begin{document}
\mainmatter              
\title{Attacking Compressed Vision Transformers}
\titlerunning{Attacking Compressed Vision Transformers}  
%
\author{Swapnil Parekh \and Pratyush Shukla \and Devansh Shah }
\authorrunning{Swapnil Parekh et al.} 
%
\tocauthor{Swapnil Parekh, Pratyush Shukla and Devansh Shah}
\institute{New York University}

\maketitle              

\begin{abstract}
Vision Transformers are increasingly embedded in industrial systems due to their superior performance, but their memory and power requirements make deploying them to edge devices a challenging task.
Hence, model compression techniques are now widely used to deploy models on edge devices as they decrease the resource requirements and make model inference very fast and efficient. But their reliability and robustness from a security perspective is another major issue in safety-critical applications. Adversarial attacks are like optical illusions for ML algorithms and they can severely impact the accuracy and reliability of models. In this work we investigate the transferability of adversarial samples across the SOTA Vision Transformer models across 3 SOTA  compressed versions and infer the effects different compression techniques have on adversarial attacks.
\end{abstract}

\section{Introduction}
Industrial systems require highly optimized algorithms which have less latency for real-time deployment. Such algorithms are neural network models that are massive in size and require high computation power. Hence deploying these models on edge devices becomes a challenging task\cite{Scaling_Vision_Transformers}. \\
ViTs are computationally expensive models with a large memory footprint so they have huge training times for massive datasets.
Model compression techniques such as quantization and pruning are now widely used to deploy such models on edge devices as they decrease the resource requirements and make model inference fast and efficient. Additionally, knowledge distillation is being used to improve model performance and memory footprints\cite{KD}.\\
More recently, ViTs have been attacked by several white boxes and black-box attacks\cite{Adversarial_Robustness_Compare}. These adversarial attacks are like optical illusions for machines, where such samples can severely impact the accuracy and reliability of models. Hence from a security perspective their reliability and robustness is a major issue in safety-critical applications.

\textbf{Novel Contribution:} We investigate the transferability of adversarial samples across the SOTA NLP models and their compressed versions and infer the effects different compression techniques have on adversarial attacks. Hence with the recent popularity of ViT-based architectures as alternative to CNNs, it is vital for the community to understand their adversarial robustness and their compressed versions. We were inspired from Yiren Zhao et. al\cite{NN_Compression} which performs an exhaustive study of white and black box attacks on CNNs and replicated them for ViTs and their SOTA compressed variants.

\section{Dataset}
The ImageNet\cite{imagenet} dataset contains 14,197,122 WordNet-annotated images. The dataset has been used in the ImageNet Large Scale Visual Recognition Challenge (ILSVRC), an image classification and object identification benchmark, since 2010. A set of manually annotated training photos is included in the publicly available dataset. A series of test photos is also available, although without the manual comments. There are two types of ILSVRC annotations: (1) image-level annotations that include a binary label for the presence or absence of an object class in the image, such as "there are cars in this image" but "there are no tigers," and (2) object-level annotations that include a tight bounding box and class label around an object instance in the image. We use the image classification type dataset description follows:
\begin{itemize}
    \item Images from 1000 different classes are included in the dataset.  
    \item It is divided into three sections: training (1.3 million images), validation (50,000 images), and testing (10,000 images) (100K images with held-out class labels).
    \item There are 14197122 photos in total.
\end{itemize}

\section{Metrics}
We use 3 metrics to quantify our attacks and model compression:
\begin{enumerate}
    \item \textbf{ASR(Attack Success Rate)} - This metric tracks how many attacks on a dataset were successful in making the model misclassify.
    \item \textbf{FLOPs} and \textbf{Throughput(Images/s)} are used to benchmark compressed model performance.
    \item \textbf{Model size (in MBs)} is used to measure compressed model memory reduction.
\end{enumerate}

\section{Vision Transformers and Types}
\subsection{Transformers}
Vision Transformers are based on Transformers \cite{Transformers} which is an attention-based sequence transduction neural network model that learns context and meaning by tracking relationships in sequential text data. The attention mechanism allows the model to make predictions by analyzing the entire input but selectively attending to some parts. Transformers apply this mechanism using an encoder-decode structure. Unlike Recurrent Neural networks (LSTMs for example), Transformers read all the words in the text as input thus parallelizing the process. This makes Transformers easily trainable on a large corpus.

\subsection{Vision Transformers}
While the Transformer architecture has emerged as the de facto standard for natural language processing tasks, its applications to computer vision remain limited. Attention is used in vision either in conjunction with convolutional networks or to replace specific components of convolutional networks while maintaining their overall structure. \cite{ViT}] demonstrates that relying on CNNs is not required and that a pure transformer applied directly to sequences of image patches can perform very well on image classification tasks. Vision Transformer (ViT) achieves excellent results compared to state-of-the-art convolutional networks when pre-trained on large amounts of data and transferred to multiple mid-sized or small image recognition benchmarks (ImageNet, CIFAR-100, VTAB, etc.).\\
In ViT each image is split into a sequence of fixed-size non-overlapping patches, which are then linearly embedded. A $[CLS]$ token is added to serve as a representation of an entire image, which can be used for classification. The authors also add absolute position embeddings that are added to the patch embeddings to retain positional information. In Computer Vision, these embeddings can represent either a 1-dimensional flattened sequence position or a 2-dimensional position of a feature. The Transformer encoder module comprises a Multi-Head Self Attention ( MSA ) layer and a Multi-Layer Perceptron (MLP) layer. The MHA layer split inputs into several heads so that each head can learn different levels of self-attention. The outputs of all the heads are then concatenated and passed through the MLP. However, the high performance of the ViT results from pre-training using a large-size dataset such as JFT-300M, and its dependence on a large dataset is interpreted as due to low locality inductive bias\cite{ViT}.

\subsection{Data-Efficient Image Transformers (DeiT)} 
DeiT presents a transformer-specific teacher-student strategy \cite{deit}. It is based on a distillation token to ensure that the student learns from the teacher through attention. This token-based distillation is gaining popularity, especially when using a convnet as a teacher. The results are competitive with convnet for both Imagenet (85.2 percent accuracy achieved) and transferring to other tasks. \\
The first essential component of DeiT is its training strategy. Initially, the authors used data augmentation, optimization, and regularization to simulate CNN training on a much larger data set. They also altered the Transformer architecture to allow for native distillation. (Distillation is the process by which one neural network (the student Neural Network) learns from the output of another (the teacher Neural Network). As a teacher model for the Transformer, a CNN is used. The use of distillation may impair neural network performance. As a result, the student model learns from two sources that may diverge: a labeled data set (strong supervision) and the teacher. To address this, a distillation token is introduced - a learned vector that flows through the network with the transformed image data, cueing the model for its distillation output, which can differ from its (distillation token's) class output. This enhanced distillation method is unique to Transformers.

\section{Attacks}
Depending on the accessibility to the target model, adversarial attacks can be divided into white box ones that require full access to the target model and query-based black-box attacks. Adversarial attacks can be divided into image-dependent ones and universal ones. Specifically, contrary to image-dependent attacks, a single perturbation, i.e. universal adversarial perturbation (UAP) exists to fool the model for most images.

\subsection{White Box Attacks}
In White Box attack\cite{Adversarial_Robustness_Compare}, the adversary has full access and knowledge of the model, that is, the architecture of the model, its parameters, gradients, and loss with respect to the input as well as possible defense mechanisms known to the attacker. It is thus not particularly difficult to attack models under this condition and common methods exploit the model’s output gradient to generate adversarial examples. Some examples are: FGSM: Fast Gradient Sign Method, FGM: Fast Gradient Method, I-FGSM: Iterative version of FGSM, MI-FGSM: Momentum Iterative Gradient-Based, PGD: Projected Gradient Descend, DeepFool: Iterative algorithm to efficiently compute perturbations that fool deep networks. 
We use Universal Adversarial Perturbations (UAP) to attack our models.

\subsubsection{Universal Adversarial Perturbations (UAP)}
UAP's goal is to identify a single little picture modification that deceives a state-of-the-art deep neural network classifier on all-natural images. Seyed-Mohsen Moosavi-Dezfooli et. al\cite{UAP} demonstrates the presence of such quasi-imperceptible universal perturbation vectors that lead to high probability of miss-classifying pictures. The label estimated by the deep neural network is modified with high probability by introducing such a quasi-imperceptible perturbation to natural pictures. Because they are picture agnostic and work on any inputs, such perturbations are named universal.\\
An example of UAPs is shown in Figure: ~\ref{fig:UAP attack example}. The authors propose the following method for estimating such perturbations - $\mu$ denotes a distribution of images in $R^d$, and $\hat{k}$ defines a classification function that outputs for each image $x \in R^d$ an estimated label $\hat{k}(x)$. The main focus of their paper is to seek perturbation vectors $v \in R^d$ that fool the classifier $\hat{k}$ on \textit{almost all} data points sampled from $\mu$. They seek a vector $v$ such that:
$ \hat{k} (x+v) \neq \hat{k} (x) \text{ for ``most'' } x \sim \mu.$

\subsection{Black Box Attacks}
\label{section:black-box-att}
In contrast to white-box attacks, a black-box attack\cite{Black_Box_Survey} has limited knowledge of the model. Querying the model on inputs and viewing the labels or confidence scores is a common paradigm for an attack with black-box limitations. While black-box assaults restrict the attacker's capabilities, they are more realistic in real-world scenarios. Security assaults are often carried out on fully developed and deployed systems. In the actual world, attacks with the goal of circumventing, disabling, or compromising integrity are common. This paradigm allows for the consideration of two parties: adversary and challenger. The challenger is the party that trains and installs a model, whereas the adversary is the party that attempts to break the system for a predetermined aim. This option allows for a variety of capability settings that mimic real-world behavior.\\
Some examples of black-box attacks are, Query Reduction using Finite Differences, Translation-Invariant, Spatial Attack, L2 Contrast Reduction Attack, Salt And Pepper Noise Attack, Linear Search Blended Uniform Noise Attack. We illustrate the attacks used in our work.

\subsubsection{Spatial Attack}
One important criterion for adversarial examples is that the perturbed images should “look similar to" the original instances. All the existing approaches directly modify pixel values, which may sometimes produce noticeable artifacts. Instead,Spatial Attack\cite{FoolBox} aims to smoothly change the geometry of the scene while keeping the original appearance, producing more perceptually realistic adversarial examples. In this attack, the idea is to use rotations and translations in an adversarial manner.\\
The attack used the following hyperparameters: \textit{do-rotations} (If False no rotations will be applied to the image), \textit{do-translations} (If False no translations will be applied to the image).

\begin{figure}[h!]
  \centering
  \begin{minipage}[b]{0.33\textwidth}
     \includegraphics[width=\textwidth]{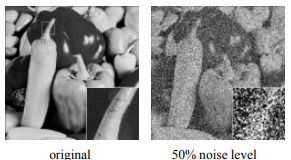}
    \caption{Salt and Pepper Noise Effects}
    \label{fig:salt_pepper}
  \end{minipage}
  \hfill
   \begin{minipage}[b]{0.3\textwidth}
    \includegraphics[width=\textwidth]{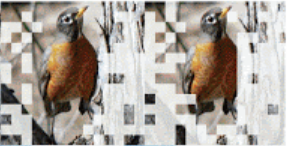}
    \caption{Pruning Example for Imagenet}
    \label{fig:pruning_example}
  \end{minipage}
  \hfill
  \begin{minipage}[b]{0.25\textwidth}
    \includegraphics[width=0.9\textwidth]{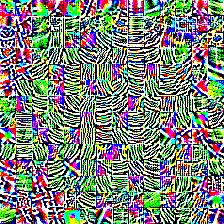}
    \caption{UAP Example}
    \label{fig:UAP attack example}
  \end{minipage}
\end{figure}

\subsubsection{Salt And Pepper Noise Attack}
Salt-and-pepper noise, also known as impulse noise, is a form of noise sometimes seen on digital images. This noise can be caused by sharp and sudden disturbances in the image signal. It presents itself as sparsely occurring white and black pixels. The results of adding this kind of noise can be observed in Figure \ref{fig:salt_pepper}. An effective noise reduction method for this type of noise is a median filter or a morphological filter. This adversarial attack\cite{FoolBox} aims to increase the amount of salt and pepper noise until the input is misclassified.\\
The attack used the following hyperparameters: \textit{steps} (The number of steps to run). 

\subsubsection{Linear Search Blended Uniform Noise Attack}
Linear Search Blended Uniform Noise Attack\cite{FoolBox} aims to blend the input with a uniform noise input until it is misclassified.\\
The attack used the following hyperparameters: \textit{distance} (Distance measure for which minimal adversarial examples are searched), \textit{directions} (Number of random directions in which the perturbation is searched), \textit{steps} (Number of blending steps between the original image and the random directions).

\section{Compression Techniques}
\subsection{Dynamic Quantization}
\label{section:comp1} 
When developing neural networks, there are several trade-offs to consider. A recurrent neural network's number of layers and parameters can be changed during model building and training, allowing you to trade off accuracy with model size and/or model latency or throughput. Because we are iterating over the model training, such adjustments might take a long time and a lot of computing resources. After training, quantization\cite{Dynamic_Quant} can be used to make a similar trade-off between performance and model correctness using a known model.\\
Quantizing a network entails transforming the weights and/or activations to a lower precision integer representation. This reduces model size and allows your CPU or GPU to do higher-throughput arithmetic operations. Converting from floating-point to integer numbers entails multiplying the floating-point value by a scaling factor and rounding the result to the nearest whole number. The various quantization methods take different techniques to obtaining that scale factor. Because dynamic quantization has few adjustment options, it is ideally suited for inclusion in production pipelines as a normal element of converting models to deployment.\\
Model parameters, on the other hand, are known during model conversion and have been transformed and saved in INT8 format ahead of time. The quantized model uses vectorized INT8 instructions for arithmetic. To avoid overflow, accumulation is usually done with INT16 or INT32. If the following layer is quantized or converted to FP32 for output, the higher precision number is scaled back to INT8.

\subsection{Pruning: Dynamic DeiT}
\label{section:comp2}
In ViTs, attention is scarce and the final prediction is only based on a subset of the most informative tokens, which is sufficient for accurate image recognition. Based on this observation, the authors propose a dynamic token sparsification framework for progressively and dynamically pruning redundant tokens based on the input\cite{Dynamic_ViTs}. A lightweight prediction module estimates the importance score of each token based on the current features. This module is added to different layers to prune redundant tokens hierarchically. The authors propose an attention masking strategy to differentiably prune a token by blocking its interactions with other tokens to optimize the prediction module from start to finish. Because of the nature of self-attention, unstructured sparse tokens are still hardware friendly, making it easy for this framework to achieve actual speed-up. \\
By pruning 66 percent of the input tokens hierarchically, it significantly reduces 31 percent 37 percent FLOPs and improves throughput by more than 40 percent while maintaining accuracy within 0.5 percent for various vision transformers. An example is displayed here Figure: ~\ref{fig:pruning_example} \\
DynamicDeiT \cite{Dynamic_ViTs} models, when equipped with the dynamic token sparsification framework, can achieve very competitive complexity/accuracy trade-offs when compared to state-of-the-art CNNs and vision transformers on ImageNet. \\
We trained 3 dynamic DeiT models pruned at 3 different pruning probabilities - 0.5, 0.6 and 0.7.

\subsection{Weight Multiplexing + Distillation: Mini-DeiT}
\label{section:comp3}
ViT models contain a large set of parameters, restricting their applicability on low memory devices. To alleviate this problem, the authors propose Mini-DeiT\cite{ViT_Weight_Multiplexing}, a new compression framework, which achieves parameter reduction in vision transformers while retaining the same performance.\\
The central idea of Mini-DeiT is to multiplex the weights of consecutive transformer blocks. More specifically, the authors make the weights shared across layers while imposing a transformation on the weights to increase diversity. Weight distillation over self-attention is also applied to transfer knowledge from large-scale ViT models to weight-multiplexed compact models.

\subsubsection{Weight Multiplexing}
Weight Multiplexing\cite{ViT_Weight_Multiplexing} combines multi-layer weights into a single weight over a shared part while involving transformation and distillation to increase parameter diversity.\\
More concretely, as shown in Figure \ref{fig:weight_multiplexing}, the weight multiplexing method consists of sharing weights across multiple transformer blocks, which can be considered as a combination process in multiplexing; introducing transformations in each layer to mimic demultiplexing; and applying knowledge distillation to increase the similarity of feature representations between the models before and after compression.
\subsubsection{Weight Distillation}
To compress the large pre-trained models and address the performance degradation issues induced by weight sharing, weight distillation\cite{ViT_Weight_Multiplexing} is used, to transfer knowledge from the large models to the small and compact models. Three types of distillation for transformer blocks, i.e., prediction-logit distillation\cite{prediction_logit_distillation}, self-attention distillation\cite{attention_distillation}, and hidden-state distillation. 

\section{Experiments and Results}
Our experiments work by creating attacks in 2 ways:
\begin{itemize}
    \item Create attacks on the original model and testing them on a compressed model. 
    \item Create attacks on the compressed model and testing them on the original model.
\end{itemize}
\begin{figure}[h!]
  \centering
  \begin{minipage}[t]{0.5\textwidth}
     \includegraphics[width=1.0\textwidth]{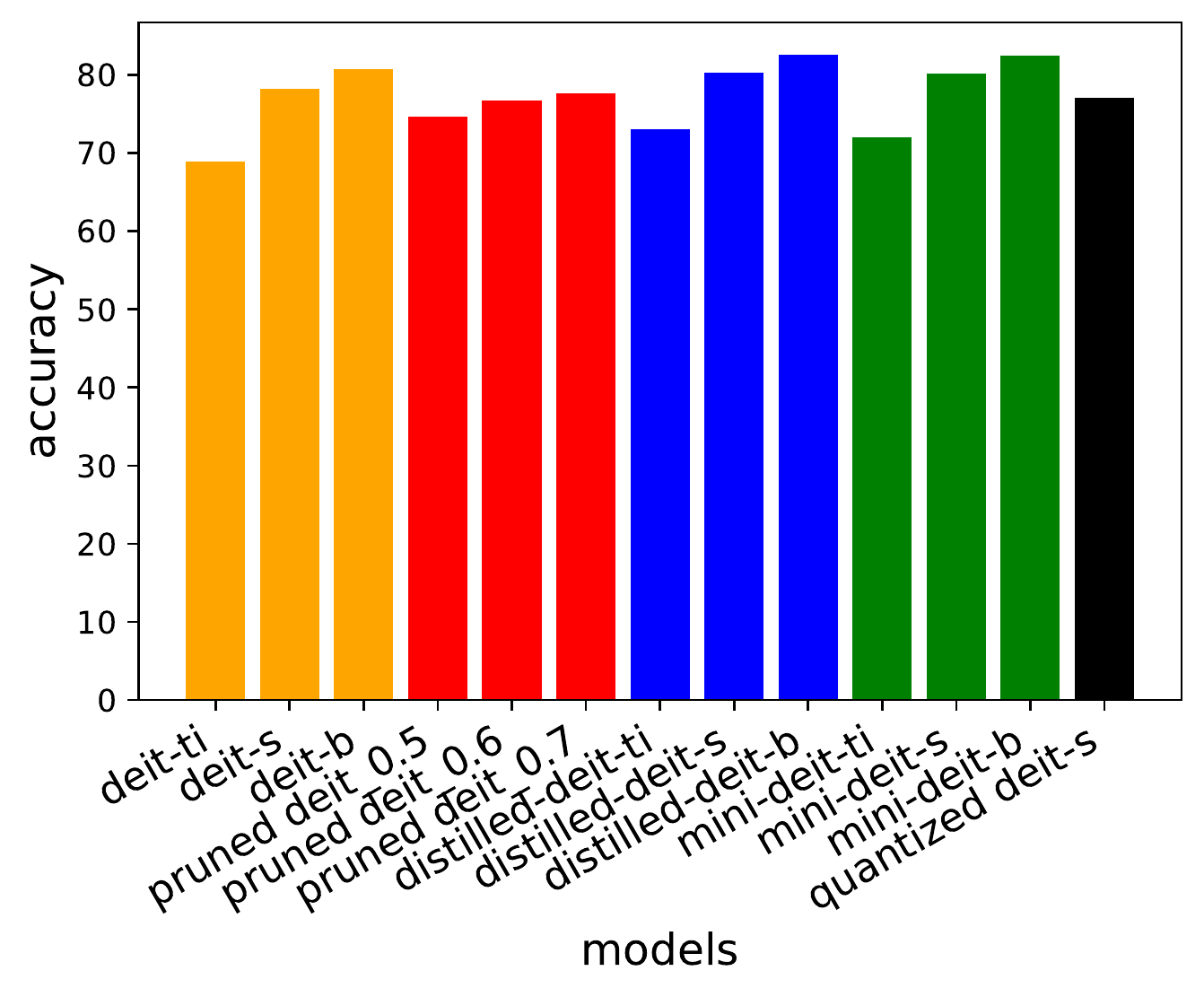}
    \caption{Accuracies of various architectures we used on the ImageNet Validation}
    \label{fig:model_accuracies}
  \end{minipage}
  \hfill
  \begin{minipage}[b]{0.45\textwidth}
    \includegraphics[width=\textwidth]{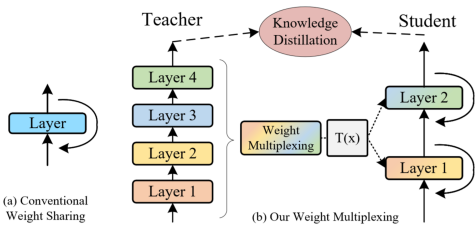}
    \caption{Weight Multiplexing Method}
    \label{fig:weight_multiplexing}
  \end{minipage}
\end{figure}
Figure ~\ref{fig:model_accuracies} shows the accuracies of our 4 models families on the Imagenet Validation set. The four families are pruned-deit, original deit, distilled deit and mini deit(which performs weight multiplexing + distillation).\\\\

We make the following observations: 
\begin{enumerate}
    \item Pruning models: As keep-rate of pruning increases and sparsity decreases the accuracy increases, and approaches the original deit small
    \item Quantized models have slightly lower accuracy compared to the deit-small
    \item Tiny $<$ Small $<$ Base accuracy for original, distilled and mini-deit
\end{enumerate}

\subsection {Compression Results}
As mentioned in sections ~\ref{section:comp1}, ~\ref{section:comp2}, ~\ref{section:comp3} we use quantization, pruning, distillation(born again networks) and weight multiplexing+distillation to compress our models. \\
The model size trend chart is shown in Figure ~\ref{fig:model-size} for the various distilled and quantized models. We can observe that born-again networks are slightly bigger than the original model, while the mini-deit size is proportional to the size of the original with a marginal drop in accuracy according to Figure ~\ref{fig:model_accuracies}. For the base model in particular mini-deit compresses deit-b from 86M to 9M(9.7x). Quantization gives the biggest model size reduction, but by doing this we lose access to gradients, forcing us to rely on less-effective black-box attacks discussed in ~\ref{sec:quant_res}.

\begin{table}[h]
\centering
\begin{tabular}{|l|l|l|l|l|}
\hline
      & main & distilled & mini-deit  & 8bit Quantization\\\hline
tiny  & 22M  & 23M       & 12M     &    6.4M                    \\\hline
small & 85M  & 86M       & 44M      &   23M                    \\\hline
base  & 331M & 334M      & 169M       &  89M                  \\\hline
\end{tabular}
\caption{Model Weight Size for Distillation + Quantization}
\label{table:model_size_table}
\end{table}
\vspace{-5mm}

In the case of the Pruning models, Table ~\ref{table:pruning_stats} shows how the FLOPS and Throughput vary as the keep-probability decreases and sparsity increases. While pruning doesn't decrease model size, it improves throughput and FLOPS which help in faster inference. 

\begin{table}[h]
\centering
\begin{tabular}{|l|l|l|l|l|}
\hline
  Metric    & main(1.0) & pruned\_0.7 & pruned\_0.6 & pruned\_0.5 \\\hline
GFLOPS      & 4.6       & 2.9         & 2.4         & 1.9 \\\hline
Throughput(img/s) & 1337.7 & 2062.1   & 2274.09     & 2526.93 \\\hline

\end{tabular}
\caption{Model Compression Performance for Pruning Models(DynamicDEiT) with different keep probability}
\label{table:pruning_stats}
\end{table}

\vspace{-10mm}
\subsection {Quantization Attack Results}
\label{sec:quant_res}

\begin{figure}[h!]
    \centering
    \includegraphics[width=10cm]{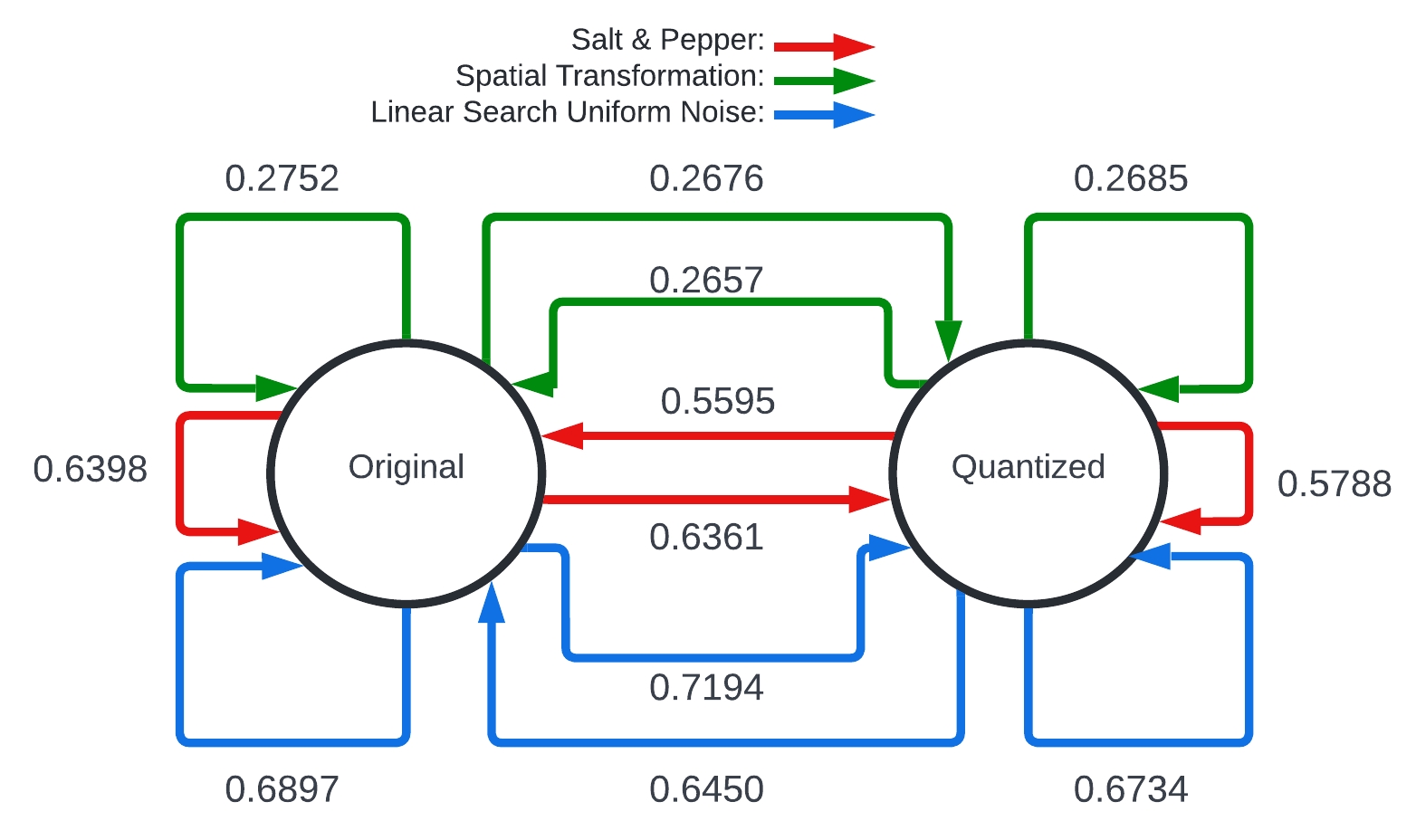}
    \caption{Results on attacking Quantized Models: The scores in each cell are the ASR of Spatial Attack| Salt and Pepper Attack| Linear Search Blended Uniform Noise Attack}
    \label{fig:quant_res}
\end{figure}

Quantized models do not have the ability "pass gradients" through them in PyTorch.
Hence we cannot use white box attacks on them, since they require access to model weights and gradients.
We use the 3 black box attacks discussed in Section ~\ref{section:black-box-att} and their ASR are reported in Table: ~\ref{fig:quant_res}.
Key Takeaways:
\begin{enumerate}
    \item ViTs are secure against Spatial Attacks.
    \item Quantized models are more vulnerable to attacks compared to the original model.
    \item Attacks from Quantized models transferred to the original model work better than vice-versa. We hypothesize this is due to the fact that 8-bit quantization is more robust creating stronger attack which work on the original model(which is still in the same feature space).
\end{enumerate}

\subsection {Pruning Attack Results}
\begin{figure}[h!]
  \centering
  \begin{minipage}[b]{0.6\textwidth}
     \includegraphics[width=1.1\textwidth]{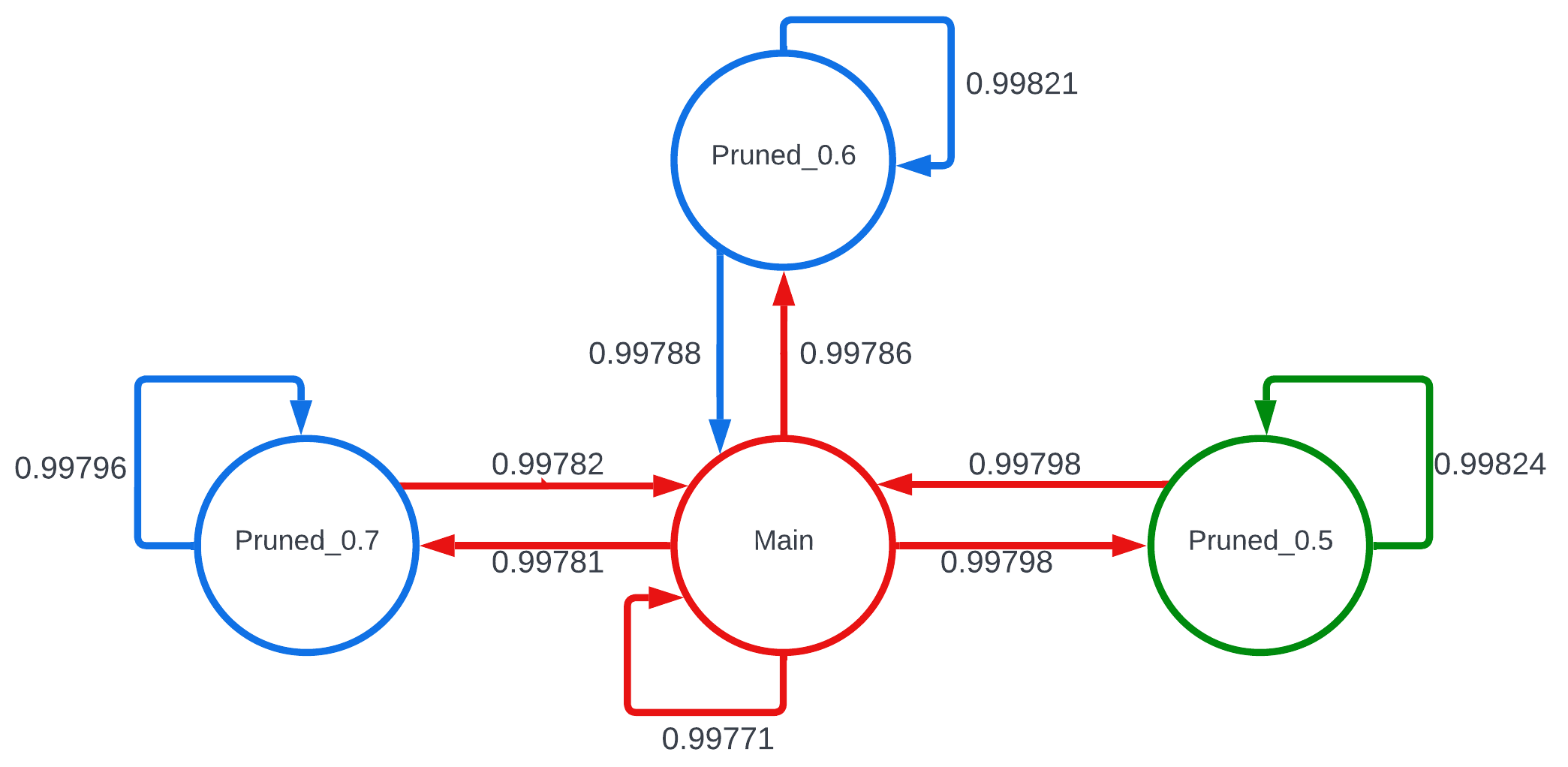}
     \vspace{-5mm}
    \caption{Pruning Results}
    \label{fig:pruning_res}
  \end{minipage}
  \hfill
  \begin{minipage}[b]{0.35\textwidth}
    \includegraphics[width=1.1\textwidth]{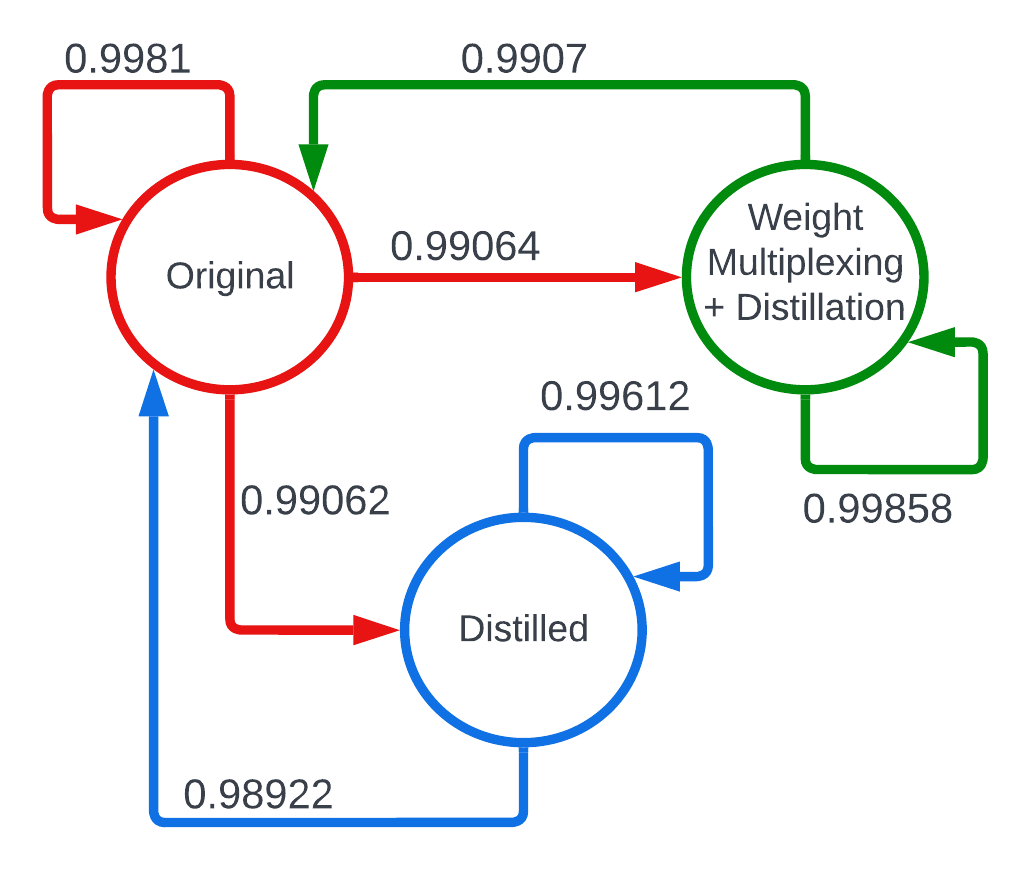}
    \vspace{-5mm}
    \caption{Base Distillation Results}
    \label{fig:base_distillation_res}
  \end{minipage}
\end{figure}

We train 3 models with varying sparsity and pruning ratios, and transfer white box attacks to the original model, results in Figure ~\ref{fig:pruning_res}.
Key Takeaways:
\begin{enumerate}
    \item Pruned models are more sensitive to attacks than the original model.
    \item The attacks remain highly transferable from the main to pruned models, which the transferability decreasing as the pruning probability decreases, which makes sense, since increasing sparsity distorts the model structure and the attacks don't transfer as well.
\end{enumerate}

\subsection {Weight Multiplexing + Distillation Attack Results}
\begin{figure}[h!]
  \centering
  \begin{minipage}[b]{0.4\textwidth}
     \includegraphics[width=1.1\textwidth]{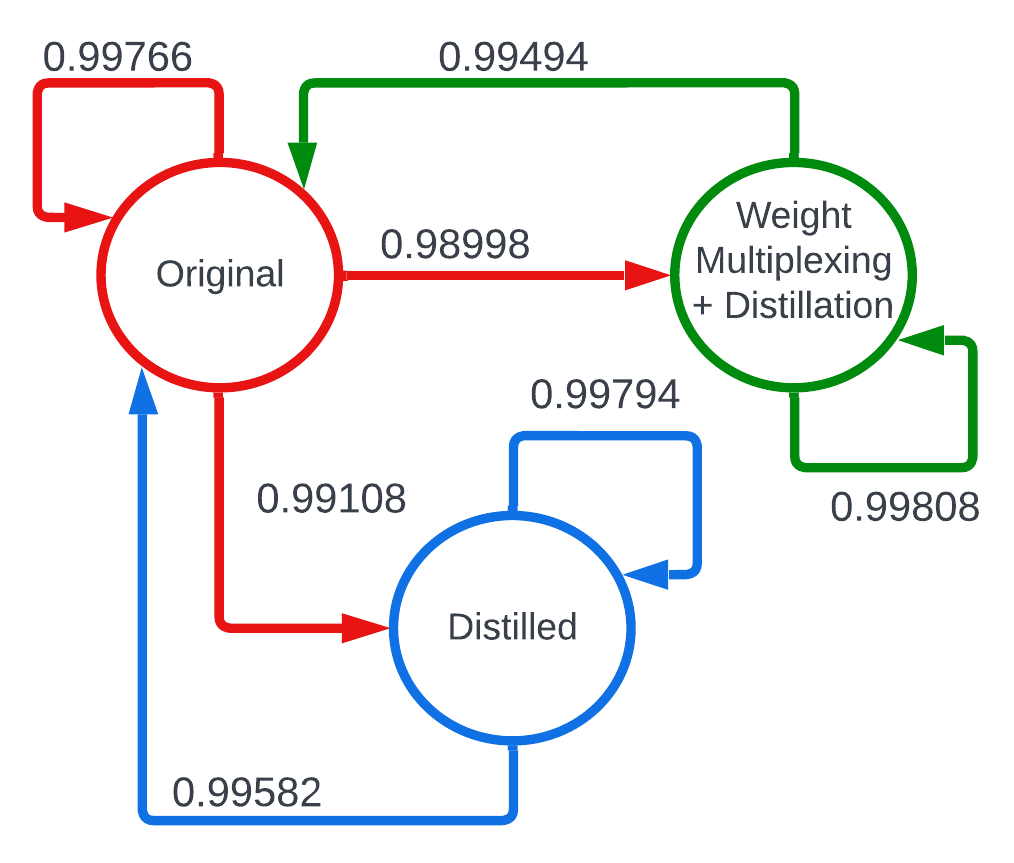}
     \vspace{-5mm}
    \caption{Tiny Distillation Results}
    \label{fig:tiny_distillation_res}
  \end{minipage}
  \hfill
  \begin{minipage}[b]{0.4\textwidth}
    \includegraphics[width=1.1\textwidth]{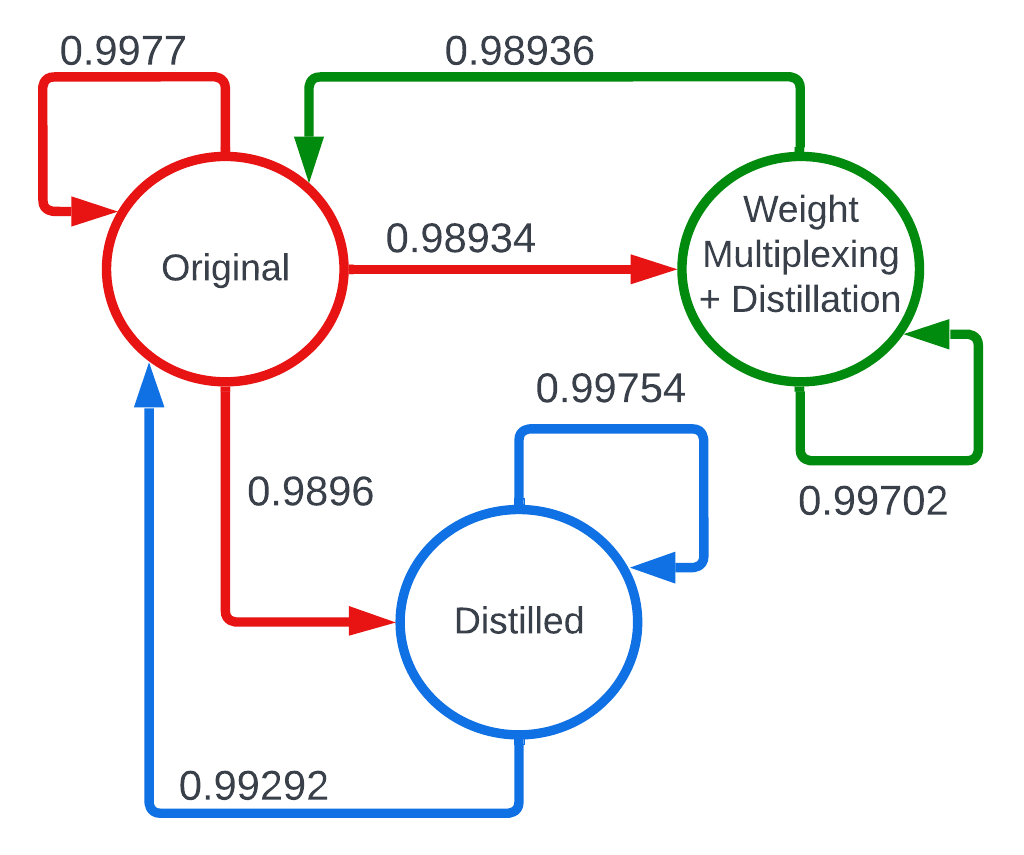}
    \vspace{-5mm}
    \caption{Small Distillation Results}
    \label{fig:small_distillation_res}
  \end{minipage}
\end{figure}

We use 2 types of distillation to see the effect that each individual components have on the attacks and their transferability for the 3 model variants, tiny, small, and base.  
Key takeaways according to Figures: ~\ref{fig:base_distillation_res},  ~\ref{fig:tiny_distillation_res},  ~\ref{fig:small_distillation_res}:
\begin{enumerate}
    \item The weight multiplexed models are more robust to attacks compared to the distilled and original models in all variants.
    \item Attacks transfer very well between from the original to both distilled models, probably because the distilled models are highly dependent on the teacher's(original model) probabilities which are used to compute the attacks in the black box setting.
    \item Base models are more robust than the Tiny and Small models variants. 
    \item Attacks from the weight multiplexed models work better than purely distilled models on the original model, which likely results from its robustness to attacks which creates stronger attacks.
\end{enumerate}

\section{Conclusion}
In this work we train various ViTs and apply state-of-the-art compression techniques on them to evaluate their robustness under an adversarial setting.\\
We discover several interesting facts such as high pruning sparsity reduces the effectiveness of adversarial attacks; compressed models, especially quantized attacks are more vulnerable to  black box attacks and weight multiplexed models are more robust to attacks compared to the others. \\
Additionally, we provide an experimentation framework to apply various compression techniques and adversarial attacks on various ViTs which is easily extensible to new compression techniques and ViTs-variants.\\
Ultimately, while compressed models may provide performance benefits, they do not provide much in way of security. The attacks created on a compressed model deployed onto an edge device can be translated to other underlying models and can wreak havoc if not dealt with promptly.\\
Our code is available here:\url{https://github.com/SwapnilDreams100/Attacking_Compressed_ViTs}

\section{Acknowledgements}
We would like to thank Prof. Siddharth Garg at NYU Tandon School of Engineering for his advice and support for this paper. 
This work was supported in part through the NYU IT High Performance Computing resources, services, and staff expertise.


\printbibliography
\end{document}